%% file: main.tex
\documentclass{mva_style}
\usepackage{graphicx}

\graphicspath{{./images/}}

\finalcopy 

\begin{document}
\title{Phenotypic Profiling of High Throughput Imaging Screens with Generic Deep Convolutional Features}


\author{
  Philip T. Jackson\textsuperscript{1}, Yinhai Wang\textsuperscript{2}, Sinead Knight\textsuperscript{2}, Hongming Chen\textsuperscript{2}, \\ Thierry Dorval\textsuperscript{2}, Martin Brown\textsuperscript{2}, Claus Bendtsen\textsuperscript{2}, Boguslaw Obara\textsuperscript{1} \\
  \and
  \textsuperscript{1}Department of Computer Science, Durham University\\
  \and
  \textsuperscript{2}IMED Biotech Unit, AstraZeneca\\ \vspace{-10pt}
  \and
  {\tt claus.bendtsen@astrazeneca.com, boguslaw.obara@durham.ac.uk}
}




\maketitle

\section*{\centering Abstract}
\textit{
  While deep learning has seen many recent applications to drug discovery, most have focused on predicting activity or toxicity directly from chemical structure. Phenotypic changes exhibited in cellular images are also indications of the mechanism of action (MoA) of chemical compounds. In this paper, we show how pre-trained convolutional image features can be used to assist scientists in discovering interesting chemical clusters for further investigation. Our method reduces the dimensionality of raw fluorescent stained  images from a high throughput imaging (HTI) screen, producing an embedding space that groups together images with similar cellular phenotypes. Running standard unsupervised  clustering on this embedding space yields a set of distinct phenotypic clusters. This allows scientists to further select and focus on interesting clusters for downstream analyses. We validate the consistency of our embedding space qualitatively with t-sne visualizations, and quantitatively by measuring embedding variance among images that are known to be similar. Results suggested the usefulness of our proposed workflow using deep learning and clustering and it can lead to robust HTI screening and compound triage.
}

\input{Chapters/introduction}
\input{Chapters/dimensionality_reduction}

\input{Chapters/clustering}
\input{Chapters/validation}
\input{Chapters/conclusion}

\bibliographystyle{plain}
\bibliography{bibliography}

\end{document}

%% file: Chapters/introduction.tex
\section{Introduction}
\label{sec:intro}

Modern drug discovery is a rapidly evolving field with the use of modern AI technology. Nonetheless, it remains expensive and time consuming for the introduction of any new medicine (almost \$3 billion, \textgreater 10 years, \cite{dimasi2016innovation}). Rather than hand-designing and testing novel drugs individually, the modern pharmaceutical approach is to use a library of compounds (e.g. 2 million compounds), and filter them with a sequence of complex imaging and biochemical tests. Typically for a high throughput screen, a single dose experiment is designed to remove vast majority of irrelevant compounds, e.g. by 100 folds.

Following the spectacular rise of deep learning techniques in computer vision, natural language processing and numerous scientific applications in recent years, deep learning has increasingly been applied to the field of chemoinformatics \cite{chen_rise_2018}. However, most recent work (e.g. \cite{bjerrum2017smiles}) has focused on predicting the biological effects of chemical compounds directly from chemical structure representations such as SMILES strings \cite{weininger1988smiles}. Morphological profiling is a complementary approach that can be used to predict a broad range of biological effects \cite{simm_repurposing_2018}. In this approach, candidate drugs are applied to cell cultures and imaged with high throughput fluorescence microscopy; depending on the bioactivity of the drugs, this can cause a variety of morphological changes to occur, yielding clues as to what effect a compound has on the cells. Indeed, \cite{simm_repurposing_2018} shows that a single network can be transferred to predict the outcomes of many other targeted assays.

Despite the rich information provided by morphological profiling, it is sometimes unknown which exact morphological phenotypes one should screen for. An example of this scenario is a high throughput screen conducted by AstraZeneca, in which novel compounds are screened for inhibition of IDOL - Inducible Degrader of the Low density lipoprotein receptor (LDLR) \cite{knight2017enabling}. In this screen, HEK293S human embryonic kidney cells were engineered to express both Low Density Lipoprotein Receptor - Green Fluorescence Protein (LDLR-GFP) and IDOL. Since IDOL degrades LDLR, the presence of green fluorescence is an indicator of IDOL inhibition. However, due to the complex and poorly understood interactions between novel compounds and human cells, presence of the GFP signal is a necessary but not sufficient condition to infer IDOL inhibition - the actual phenotypic appearance of genuine hits is not known at the screening stage. In situations like this, the expert knowledge and intuition of a biologist is required to identify phenotypes that are indicative of genuine hits. Due to the high volume of images in a HTI screen, this cannot be done manually for each individual image, and due to the unknown nature of the target phenotype, supervised learning is not applicable.

In this paper, we propose a novel procedure for computing feature vectors for cellular images using a pre-trained convolutional neural network (CNN). The resulting feature vector space can then be partitioned by unsupervised clustering, allowing us to decompose a HTI screen into a small set of visually distinct phenotypes. The expert judgement of biologists can then be applied to whole phenotype clusters rather than individual images, allowing a HTI screen to be filtered rapidly for interesting compounds. The feature vectors can also be embedded in 2D space for visualization using dimensionality reduction techniques such as t-sne \cite{maaten2008visualizing}, providing a visual summary of the phenotypic distribution.

In Section~\ref{sec:dimensionality_reduction} we describe our feature extraction pipeline. Section~\ref{sec:clustering} discusses clustering and in Section~\ref{sec:validation} we demonstrate and evaluate our approach on a high throughput imaging dataset (IDOL).

%% file: Chapters/dimensionality_reduction.tex
\section{Feature Extraction}
\label{sec:dimensionality_reduction}


CNNs trained on large datasets such as ImageNet have been found to learn a hierarchy of features, with early layers learning general, task-agnostic features pertaining to texture and shape primitives, and later layers learning more task specific features \cite{yonsinski2014transferable}. Despite the obvious differences between ImageNet images (which are generally photographs) and fluorescence micrographs, the early convolutional layers of a CNN trained on ImageNet are general enough to respond to the differences in shape, colour and texture in fluorescent labeled cellular images. 

\begin{figure}
    \includegraphics[width=\linewidth]{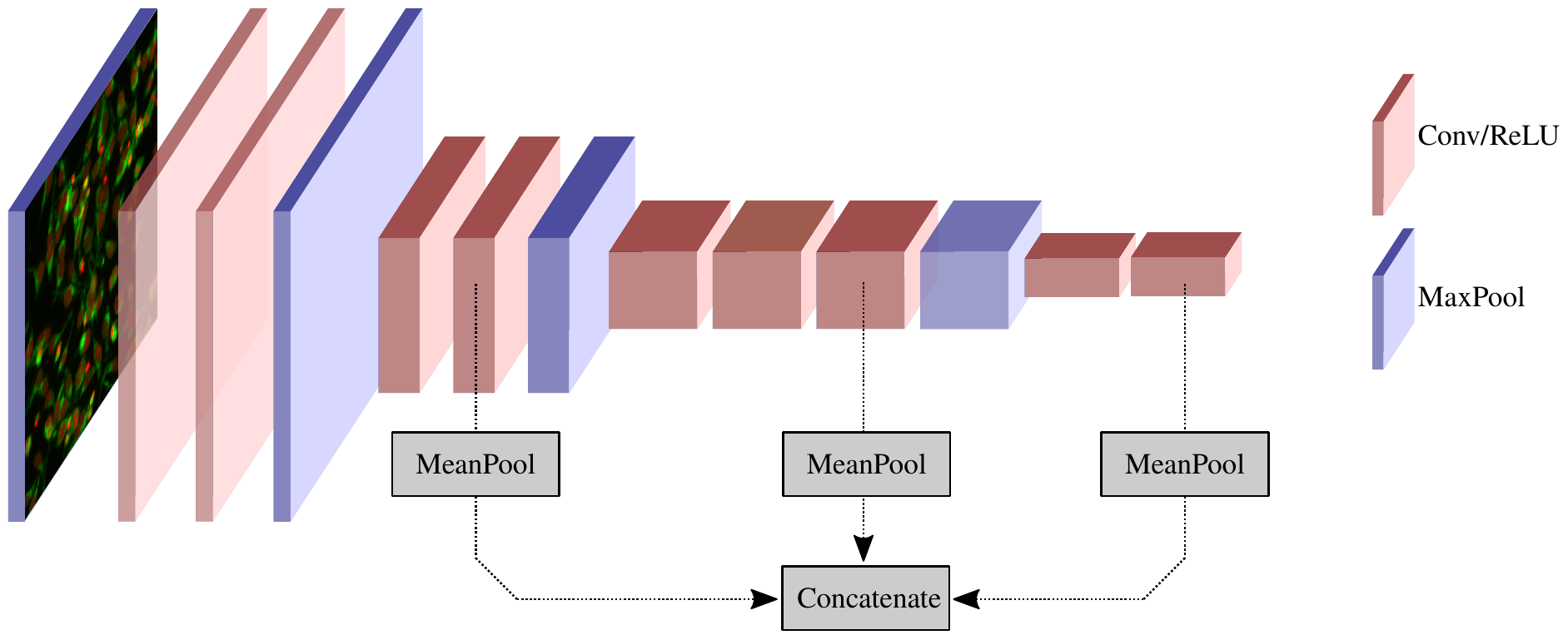}
    \caption{Our feature extraction pipeline. We feed our images through a pre-trained VGG16 network, truncated before the fully connected layers, and concatenate the spatial means of three intermediate convolutional layer activations. This yields a vector of multi-scale convolutional features, which we later embed in 2D space via t-sne (see Figure~\ref{fig:tsne}).}
    \label{fig:extraction}
\end{figure}

Our feature extraction begins by computing feature maps for a cellular image, by feeding it through a pre-trained CNN (Figure~\ref{fig:extraction}). The CNN architecture we use is VGG16 \cite{simonyan2014deep}, pre-trained on ImageNet. Rather than taking final fully-connected layer activations as our feature vector, we extract our features by mean pooling the (pre-activation) feature maps of three early convolutional layers, and concatenating the means to produce a feature vector of length $1024$ (one component for each feature map in the chosen layers). Extracting features from early convolutional layers rather than final fully connected layers has a number of advantages for large images in which the objects of interest are relatively small and homogeneously distributed.

Firstly, high level representations learned by CNNs contain information that is immediately relevant for identifying the classes they are trained to recognize. In the case of ImageNet, these are everyday objects such vehicles and animals. High level features are unlikely to be very descriptive for cellular images, which differ substantially from the ImageNet images and training classes, but lower level features are still general enough to capture information about shape, texture and colour.

Secondly, because of the fixed weight matrix connecting the first fully connected layer to the last convolutional layer, fully connected layers require the input image to be of a fixed size. By using only the convolutional layers of the network, we avoid the need to downsample our images to $224 \times 224$ (for the VGG network), which would discard valuable high frequency information.

\begin{figure}
    \centering
    \includegraphics[width=0.5\linewidth]{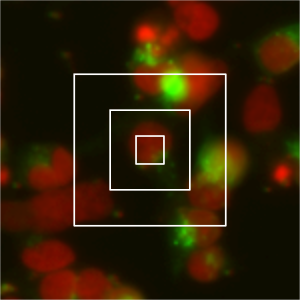}
    \caption{Receptive field sizes of our chosen convolutional layers, overlaid on a histological image for reference. By pooling from multiple layers, we can extract information about fine texture, individual cells and small clusters of cells.}
    \label{fig:receptive_fields}
\end{figure}

Thirdly, different layers capture information at different scales. Higher level layers have larger receptive fields, and describe patterns of greater size and complexity, but successively discard the higher frequency information captured by lower layers. By extracting features from the fourth, seventh and ninth convolutional layers, we obtain a multi-scale representation (see Figure~\ref{fig:receptive_fields}).

%% file: Chapters/clustering.tex
\section{Clustering}
\label{sec:clustering}

\begin{figure}
    \centering
    \includegraphics[width=\linewidth]{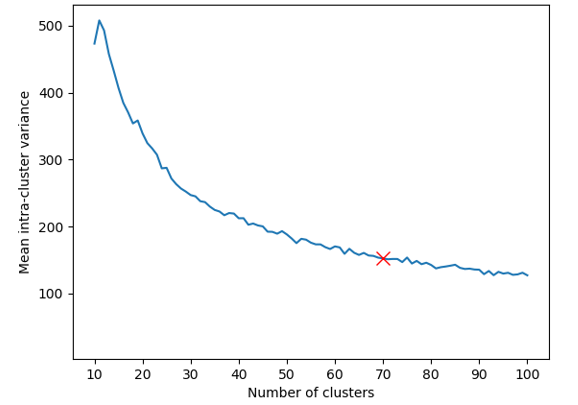}
    \caption{Mean intra-cluster variance as a function of the number of clusters. We cap the number of clusters at $70$, as diminishing returns are observed past this point.}
    \label{fig:elbow}
\end{figure}

To discover distinct phenotypes in the dataset, we perform k-means clustering in feature space. The optimal choice for the number of clusters $k$ is a trade-off between making the clusters as homogeneous as possible, and keeping their number low. Figure \ref{fig:elbow} shows the mean intra-cluster embedding variance as a function of $k$; since we observe diminishing returns past $k=70$, we choose $70$ as the optimal number of clusters.

%% file: Chapters/validation.tex
\section{Results}
\label{sec:validation}

\begin{figure}[t]
    \centering
    \includegraphics[width=0.8\linewidth]{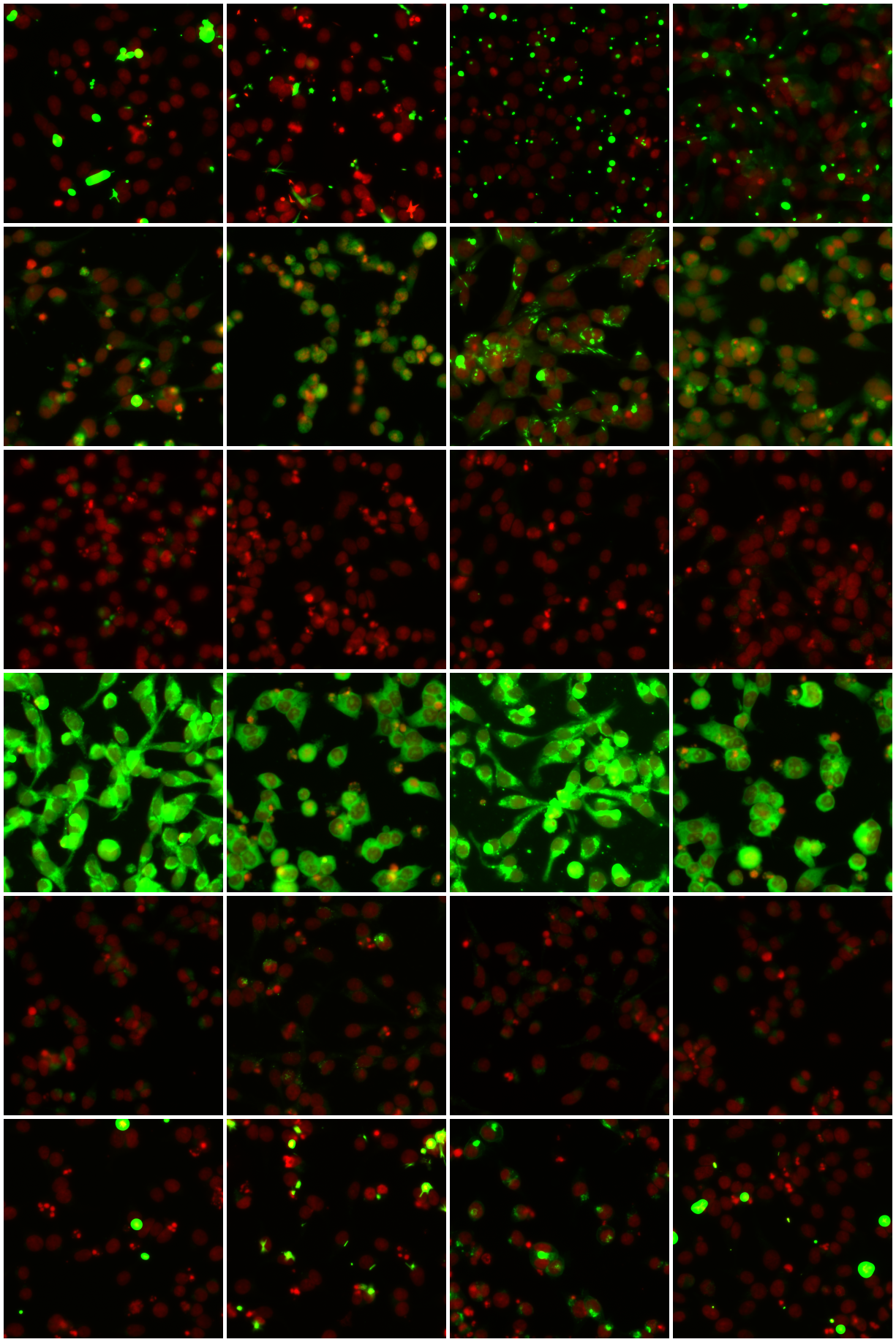}
    \caption{Samples from six of the $70$ phenotypic clusters detected by k-means. Each row shows four example images from a single cluster. Rows 2 and 4 show genuine GFP expression.}
    \label{fig:cluster_swatch}
\end{figure}

The most crucial aspect to validate is that our feature extraction does indeed embed similar images at similar points in the feature space. We can evaluate this quantitatively, by measuring the variance (concretely, mean squared distance to centroid) of different groups of image embeddings. The compounds tested in this dataset are chemically clustered by AstraZeneca in Extended-Connectivity Fingerprint embedding space \cite{rogers_extended_2010}, resulting in $711$ chemical clusters. We would expect images corresponding to compounds from the same chemical cluster to have more tightly clustered feature vectors than the dataset as a whole, because similar compounds may lead to similar morphological changes in images. As expected, the mean intra-cluster feature variance is $66.5\%$ that of the dataset as a whole. Furthermore, we would expect different images captured from the same well on an assay plate to be clustered more tightly still, because these images all correspond to the same compound. We observe the mean intra-well variance (4 images were captured per well) to be $4.0\%$ that of the full dataset.

To validate the quality of our clusters, we display samples from six phenotypic clusters in Figure~\ref{fig:cluster_swatch}. We see that k-means has identified relatively homogeneous clusters, further validating the quality of our embeddings.

\begin{figure*}[t]
    \centering
    \includegraphics[width=\linewidth]{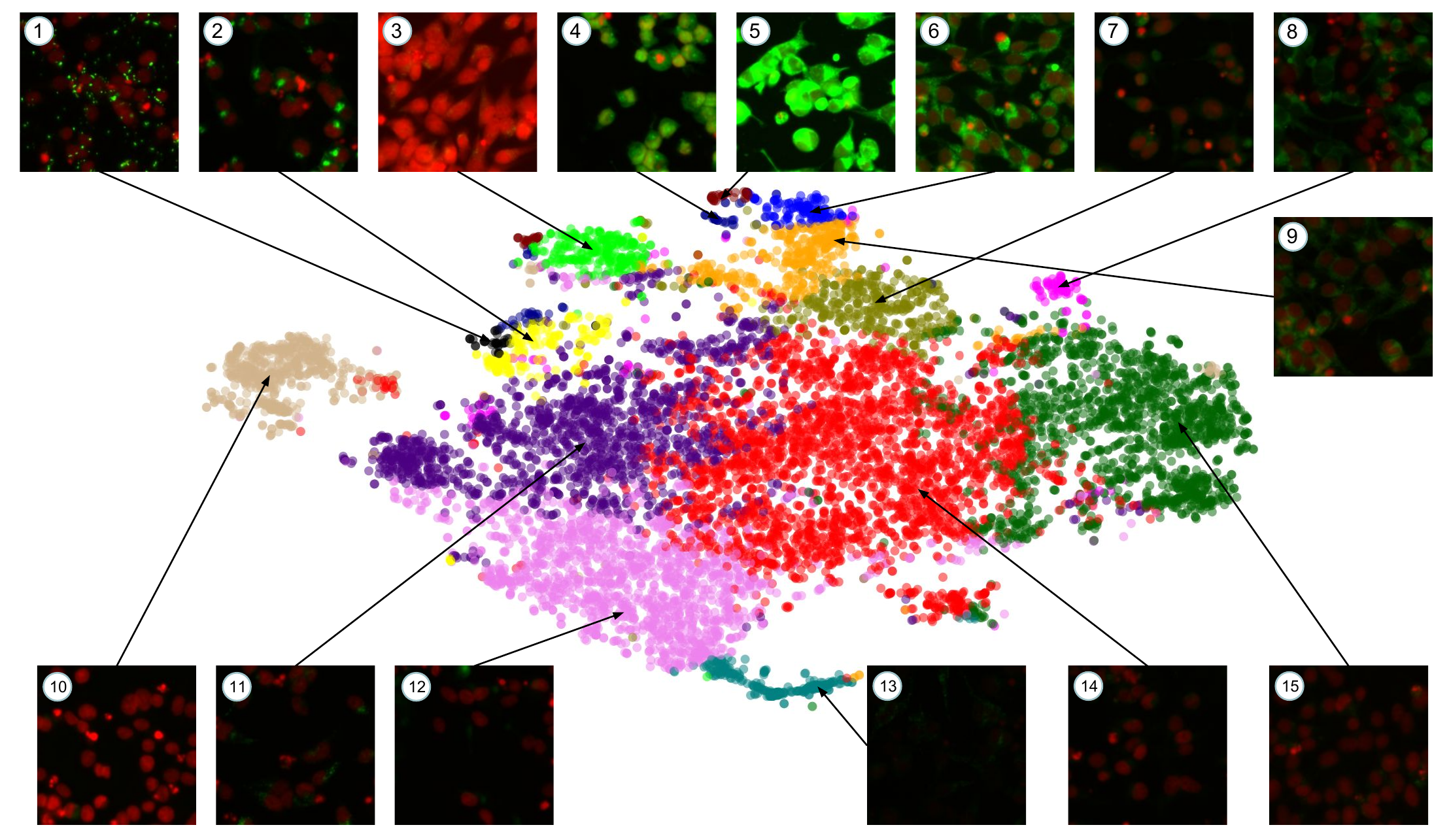}
    \caption{A t-sne embedding of our dataset, with colours showing phenotypic clusters discovered by k-means. For visualization purposes, we set $k=15$ here.}
    \label{fig:tsne}
\end{figure*}

Figure~\ref{fig:tsne} shows a t-sne visualization of our entire dataset, with phenotypic clusters detected by k-means labelled by colour and annotated with representative samples from that cluster. To speed up the t-sne process, we used Principal Component Analysis (PCA) to reduce the dimensionality of our features from $1024$ to $10$. These $10$ principal components explain $>90\%$ of the variance in the embedding space. Clusters $4$ and $6$ in Figure~\ref{fig:tsne} show genuine GFP expression, while the others are judged as uninteresting by the biologists, and can be discarded, resulting in a 47-fold reduction. Using the full $70$ clusters would result in more precise filtering and greater reduction still.

%% file: Chapters/conclusion.tex
\section{Conclusion}
\label{sec:conclusion}

We have developed a novel workflow for high throughput screening. In this workflow, images were represented as deep learning feature vectors from a pre-trained convolutional neural network. This was followed by the clustering of images with similar image phenotypes. This facilitates scientists to select interesting clusters for downstream screening in an attempt to find hit compounds. Because it uses generic convolutional features extracted from a pre-trained convolutional neural network, our method requires no training and can be applied to any cellular screen dataset without hyperparameter tuning - a significant saving in time. Our visualizations allow scientists to quickly assess the distribution of cellular morphologies in a high throughput imaging screen, or within a smaller subset of compounds, such as a chemical cluster. Meanwhile, our proposed workflow allows scientists to select interesting/promising phenotypes and quickly retrieve chemical clusters that show a high prevalence of said phenotypes.

Future work could gain further insight into the biological processes at work by investigating the relationship between our morphological embeddings and the ECFP embeddings of the chemicals that produced them, perhaps by predicting ECFP embeddings from morphological embeddings using supervised learning. Our techniques could also be applied to other imaging modalities, such as tissue pathology and mass spectrometry imaging, with minimal modification needed.